\documentclass[11pt]{article}

\usepackage[preprint]{acl}

\usepackage{times}
\usepackage{latexsym}

\usepackage{enumitem}

\usepackage{comment}

\usepackage[T1]{fontenc}

\usepackage[utf8]{inputenc}
\usepackage{natbib}
\usepackage{microtype}

\usepackage{inconsolata}

\usepackage{graphicx}
\usepackage{multirow}
\usepackage{multicol}
\usepackage{xcolor}
\usepackage{soul}
\usepackage{amssymb}
\newcommand{\giora}{\textcolor{red}}

\usepackage[table]{xcolor}

\title{Quality-Conditioned Agreement in Automated Short Answer Scoring: \\ 
Mid-Range Degradation and the Impact of Task-Specific Adaptation}

\author{Abigail Victoria Gurin Schleifer$^1$ \ \   Moriah Ariely$^1$ \ \ Beata Beigman Klebanov$^2$ \vspace{1mm}\\ 
\textbf{Asaf Salman}$^1$ \ \  \textbf{Giora Alexandron}$^1$ \vspace{1mm}\\
  $^1$ Weizmann Institute of Science, Rehovot, Israel \\
  $^2$ ETS, Princeton, USA \\
  \texttt{\{abigail.gurin-schleifer,moriah.ariely,giora.alexandron, asaf.salman\}}\\
  \texttt{@weizmann.ac.il}\\
  \texttt{bbeigmanklebanov@ets.org}\\
}

\begin{document}
\maketitle
\begin{abstract}
Automated short answer scoring (ASAS) is shifting from discriminative, fine-tuned models to large language models (LLMs) used in few-shot settings. This paradigm leverages LLMs’ broad world knowledge and ease of deployment, but limited task-specific data may reduce alignment on complex scoring tasks. In particular, its impact on scoring partially correct responses that require nuanced interpretation remains underexplored.
We investigate the relationship between the degree of task-specific adaptation of different models and quality-conditioned scoring agreement. We compare three LLMs (GPT-5.2, GPT-4o, Claude Opus 4.5) in few-shot mode, a fine-tuned BERT-based encoder, and a human expert on two open-ended biology items, using several hundred student responses and ground truth scores provided by a biology education expert.
The results show that human–human agreement is highest and stable across the full quality spectrum. All AI models perform well on fully correct and fully incorrect responses, but exhibit substantial degradation on mid-range responses. This mid-range degradation is  conditioned on task-specific adaptation: It is most severe in few-shot LLMs with few examples and decreases as task-specific data increases, with fine-tuned encoder models performing best.
This mid-range degradation may lead to inequitable evaluation of responses produced by students with developing understanding. Our findings highlight the importance of quality-conditioned fairness, with particular attention to mid-range responses.
\end{abstract}

\section{Introduction}
Automated short answer scoring (ASAS) of open-ended responses is a central application of natural language processing (NLP) in education \cite{bonthu2021automated, haller2022survey}. ASAS approaches can be classified into reference-based~–~scoring based on similarity to reference graded responses~--~and example-based, where the model learns to map student responses to rubric scores using task-specific labeled data. The fundamental trade-off is that reference-based approaches require only a few scored examples but their accuracy is typically very limited \cite{bexte2023similarity}, while example-based ones can reliably learn to mimic expert grading but require ample training data \cite{schleifer2023transformer} and are still prone to gaming \cite{ding-etal-2020-dont}.

Early example-based systems relied on handcrafted linguistic features and machine learning models to approximate human scoring \cite{haller2022survey}. More recent work has leveraged neural machine learning that can capture deeper semantic relationships in student responses. 
Such neural-based automated assessment systems primarily relied on discriminative machine learning implemented on top of encoders that transferred raw responses into vectorized embeddings \cite{condor2020exploring,li2021semantic,schleifer2023transformer}. 

The emergence of generative LLMs introduced a new paradigm for ASAS. Decoder-based language models can evaluate student responses based on their extensive pretraining and general world knowledge, relying on instructions  to guide in-context interpretation (e.g., according to given rubrics), rather than on adapting their parameters using task-specific datasets \cite{lin2023unlocking}. 

However, because generative models are not explicitly trained on the application of a specific assessment framework to student responses, their scoring may be less aligned with expert grading, particularly in assessment contexts that have domain-specific standards and require subtle interpretation \cite{wu2025unveiling, yacobson2026chemxai}. Few-shot prompting, which is a prompting strategy that provides a few labeled examples, was developed to help the model align its interpretation with that of experts \cite{lin2023unlocking}. This raises the question of the amount of task-specific data that various types of architectures require to align with expert grading. 

The alignment of automated scoring systems with human grading is typically evaluated using metrics like $\kappa$ or Pearson's correlation  between model predictions and human raters \cite{klebanov2022automated,williamson2012framework}. That said, it is also recognized that overall scoring accuracy may not be sufficient for validating an automated scoring system, as its performance can vary systematically across student sub-populations, raising fairness concerns \cite{williamson2012framework}. While the psychometric theory discusses sub-populations to be examined quite generally, e.g., as ``identifiable and relevant'' subgroups \cite{xi2010we},  
much of the literature on fairness in educational AI tends to focus on demographics-based sub-groups \cite{loukina2019many, madaio2022beyond} and on high-stakes, large-scale assessment \cite{johnson2023evaluating}. Discussing fairness in the formative classroom  assessment specifically,  \citet{camilli2006test} noted that the use of the assessment is generally ``to `locate' the presenting proficiencies
of students in order to guide instruction along the contours
of their strengths and weaknesses'', and help instructors ``see if there is anything that, if not addressed, sets certain students up to fail or otherwise miss significant opportunities to learn'' (p. 248). To help provide the right learning opportunities for learners at different current states of knowledge, these states of knowledge need to be detected accurately and equitably, as otherwise students whose knowledge state tends to be mis-recognized would be in danger of being provided inappropriate learning support. 



In the recent literature on automated analysis of student responses, there are indications that states of partial knowledge may be harder to detect automatically than completely correct knowledge states \cite{gurin2025uncovering,PhysRevPhysEducRes.19.020163} or than both completely correct and completely incorrect ones \cite{grevisse2024llm}.
 In such cases, students demonstrating partial mastery of a concept (mid-range responses) may receive less reliable evaluation than those whose responses are clearly correct or clearly incorrect. We refer as ``quality-conditioned fairness'' to the requirement that ASAS system performs equally well across the entire range of response quality. 

Due to the nuanced interpretation required for mid-range responses, we expect that task-specific labeled data will be particularly important for AI grading systems when grading such answers, and that its quantity will be associated with the systems’ accuracy on these responses.

However, little is known about agreement patterns across the response quality spectrum, and how this relates to model architectures -- encoder-based discriminative models and decoder-based generative ones -- and across different levels of task-specific adaptation (few-shot with varying levels of `few', and fine-tuning). Thus, our research is guided by the following research question:\vspace{2mm}


\noindent\textbf{RQ:} \textit{What is the association between response quality and inter-rater agreement for human-human vs human-AI pairings  for different levels of task-specific adaptation of the AI models? (few-shot with different numbers of examples vs fine-tuned)?} 
\vspace{2mm}

The data used for this study included responses from about 800 high-school students to two open questions in biology requiring students to provide scientific explanations in the domain of cellular respiration. The responses were graded according to an analytic grading rubric by two biology education experts. To answer the RQs, we compared three generative decoder-based models in a variety of few-shot modes, a discriminative fine-tuned encoder model and a human grader in terms of their agreement with the expert human grader overall and across different levels of response quality. 

Previewing the results, we show that disagreements between human and AI raters concentrate among mid-quality responses, while agreement remains high at the extremes of the quality spectrum. They also demonstrate a consistent pattern of {\bf mid-range degradation of AI scoring accuracy}, meaning that AI graders, but not the second human expert grader, tended to be less aligned with the expert on the mid-range responses, and that this tendency was partially mitigated as the AI system received more task-specific data. 


These findings highlight the importance of task-specific adaptation for accurately evaluating partially correct responses and suggest that evaluation frameworks for AI-assisted assessment should explicitly consider quality-conditioned, and specifically mid-range, assessment fairness.

\section{Related Work}


Fine-tuning the parameters of the model on task-specific data is a common technique for adapting encoder-based pre-trained language models such as BERT \cite{devlin2018bert} to the given task. The more recent models, while still utilizing the Transformer architecture, use orders of magnitude more parameters (hence their designation as {\em large} language models, or LLMs), and demonstrate in-context learning, namely, the ability to perform well on tasks that the model has not been trained for with only a small number of examples (few-shot). While larger models generally demonstrate stronger in-context learning, empirical studies suggest that parameters other than size, such as the design of the prompt, the model  architecture, the diversity of the training data, and the selection of the few-shot examples can impact the model's in-context learning ability \cite{berti2025emergent}.  




A number of studies compared the performance of fined-tuned encoders and few-shot LLMs on ASAS. \citet{chamieh-etal-2024-llms} evaluated ASAS performance on datasets from various domains, including mathematics, science, and English language arts. They found that fine-tuned models generally performed better, especially for tasks that require more complicated reasoning or domain-specific knowledge. \citet{kortemeyer2024performance} compared the performance of GPT-4, BERT, and RoBERTa‑large \cite{liu2019roberta} on ASAS datasets in varied science domains. They found that the fine-tuned BERT and RoBERTa-large outperform the general-purpose GPT-4 LLM. 
\citet{henkel2024can} found that few-shot models outperformed zero-shot ones in the context of ASAS, suggesting that more task-specific information may help improve scoring performance. \citet{ferreira2025automatic} studied ASAS in English in the context of undergraduate computer science and in Brazilian Portuguese in 8th grade biology and found that fine-tuned BERT models outperformed few-shot GPT-4o on both datasets. Across these studies, the evaluation of performance was done using overall summary measures, such as RMSE, F1, Pearson's correlation, or kappa-family metrics. In particular, it remains unclear {\em what kinds of responses} were easier or harder for the different models to score. 

There are indications in the literature on automated scoring of constructed responses that point towards a specific weakness of LLM-based scoring, namely, the tendency of these models to over-score low quality responses. \citet{chang2024automatic} examined the quality of GPT-4 scoring, in zero-shot and one-shot setting, of short answers in a variety of undergraduate courses in Finnish. They found that the models were too lenient, both in binary pass/fail scoring and in scoring on a scale: Too few responses were marked as fail, and too few responses were assigned to the bottom half of the scale. \citet{PhysRevPhysEducRes.19.020163} reported similar results for scoring short answers in introductory physics -- GPT-4 was a more lenient rater. The author noted that agreement between human and AI scores was stronger at the high end of the response quality spectrum. 
In the context of medical undergraduate courses, \citet{grevisse2024llm} observed that different LLMs were more or less severe than human raters on data from different courses. They further observed that there was stronger agreement between the LLM and human ratings for the completely incorrect and completely correct answers than for partially correct answers and noted GPT-4's especially high precision on detecting the fully correct answers.  
 These findings suggest that automated scoring should be examined from the point of view of response quality, as there is evidence that the scoring accuracy may be uneven across the score distribution. However, it is not clear whether few-shot LLM scoring is more or less susceptible to uneven performance than the more extensive task adaptation through fine-tuning. It is also not clear whether any unevenness in AI scoring may have to do with inherent difficulty of discriminating between scores at certain levels, namely, whether human scoring exhibits a similar pattern. Our contribution is a systematic examination of multiple scoring mechanisms -- human, fine-tuned pre-trained language models, and few-shot LLMs -- in terms of the relationship between scoring accuracy and response quality, in the context of ASAS.    

\section{Methodology }

\subsection{Assessment Items and Scoring Rubric}
We analyzed student responses to two open-ended items in biology.
The items deal with cellular respiration, and specifically, the effect of smoking and anemia on physical exercise. The items are conceptually similar and differ in context and surface features, and students' explanations are expected to have the same structure, and use similar argumentative flow. They are similar to questions that frequently appear on the biology national matriculation exam. They were administered in Hebrew and their translation is presented below. 

\noindent Smoking (Item 1): Cigarette smoke contains harmful substances, including CO, which binds haemoglobin more strongly than oxygen. Explain how high CO levels impair exercise ability.

\noindent  Anemia (Item 2): A man with low red blood cell levels (anaemia) reports weakness and difficulty exercising. Explain how reduced red blood cells make exercise difficult.

Due to their conceptual similarity, the items share the same grading rubric. It is an analytic grading rubric that decomposes a student's scientific explanation into a collection of 10 binary categories, each representing an essential property that the student response should include. The items and the rubric were developed by the 2nd author of this paper. The full description of the grading rubric can be found in \citet{ariely2025reflective}.

\subsubsection{Data Collection, Scoring, and Partitioning}
Student responses were collected in two cycles conducted in two different academic years, with a two-year gap between the cycles. In the first cycle, student responses to an instrument that contained both items were collected from 669 students in grades 10-12 attending 25 high schools across the country. 
For each category in the rubric, a student's response was graded as '$1$' if that category was addressed in the text, and '$0$' otherwise. Approximately $5\%$ of the 669 responses were annotated jointly by two human raters -- Rater1 (the second author) and Rater2. Then, an additional $20\%$ of the 669 responses were annotated independently by both raters
establishing inter-rater agreement on $n=130$ responses ($\kappa$ 
ranged between $0.89$ and $0.98$ across categories). Last,  
Rater1 coded the remaining responses; see \citet{ariely2022machine} for more details on the annotation process. The data was used as \textbf{training data} for the models, namely, for fine-tuning the BERT-based models and for picking the few-shot examples for the LLM-based grading.

In the second cycle, responses to the same instrument were collected from 152 students (in total, 304 responses) of similar demographics, and scored by Rater1 (the 2nd author of this paper). We refer to this scoring as the ground truth labels. These responses were used as \textbf{test data} for all the experiments described below. A second human rating was obtained for the test items as part of the current study of human vs AI raters and will be described in more detail in section \ref{sec: scoring_process}.

\subsection{Evaluation Metric for Scoring Differences}
The scoring of each response is represented as a 10-dimensional binary vector, where each dimension corresponds to a category, with ‘1’ indicating a correct response on that category. As noted above, we referred to the scoring of Rater1 as the ground truth labels. The overall score of a response is thus defined as the number of categories (i.e., bins) that Rater1 marked as correct, which ranges from 0 to 10, meaning that higher scores indicate higher response quality.

\subsubsection{Manhattan Distance for Scoring Differences}\label{sec: manhattan}
For a student response $s$, we denote the gold labels of $s$ by: $l^G_s = (l^G_1,...,l^G_{10})$, and the labels assigned by Rater2  (Human or AI-based) -- by: $l^{R_2}_s = (l^{R_2}_1,...,l^{R_2}_{10})$. $l^G_s$ and $l^{R_2}_s$ are binary $10$-dimensional vectors.
We measure the absolute difference between the two scorings by the Manhattan-distance (also called the $L_1$-distance) \cite{chen2004marriage} between $l^G_s$ and $l^{R_2}_s$:
\begin{equation}
\left\|l^G-l^{R_2}\right\|_{1} = \left|l^G_1-l^{R_2}_1\right|+...+\left|l^G_{10}-l^{R_2}_{10}\right|,
\label{eq: L1}
\end{equation}
which equals the number of categories scored differently by Rater2 compared to the gold labels. 

\subsection{Experimental Set-Up}
\subsubsection{The Raters and the Scoring Task}
The experimental set-up included comparing the alignment to the ground truth labels of the following scoring mechanisms: (i) a second human expert, (ii) a fine-tuned BERT-based model, and (iii) three LLMs -- GPT-5.2 \cite{Aaditya2025gpt}, GPT-4o \cite{achiam2023gpt}, and  Claude-opus-4.5\cite{anthropic2025}. At the time of writing, GPT-5.2 and  Claude-opus-4.5 are the frontier models of OpenAI and Anthropic, while GPT-4o represents OpenAI's previous generation model, which is the current workhorse for many LLM tasks due to its good combination of capabilities, speed, and cost. 

The task of the graders was to score each response of the test set ($152$ responses for each item) according to each of the categories of the rubric, producing a 10-dimensional binary vector per response.
For the LLMs we took the majority result:
Each LLM was invoked three times for every category on every response; the result that appeared most of the time is considered the final LLM score.

We then computed the Manhattan distance, per response and rater, between the Rater2's scoring vector and the ground truth vector, as described in Section \ref{sec: manhattan}.
 Below, we provide more details about the scoring procedure.

\subsubsection{Scoring Procedures by Scoring Mechanism}\label{sec: scoring_process}
Below we describe how the scoring of the test set was implemented for each scoring mechanism. 

\begin{enumerate}[leftmargin=*, labelindent=0pt, nosep]

\item \textbf{Human}: The test set was labeled according to the rubric by a second biology education expert who scored part of the 1st cycle dataset (that was used as training data in this study) a few years ago, after a short recap with Rater1 on the rubric and its implementation.

\item \textbf{Fine-tuned BERT-based Classifier}: 
The BERT-based model DictaBERT 
\cite{shmidman2023dictabert} was fine-tuned on the training data (n = 669).
A separate binary classifier was trained for each item and category, meaning that overall, 20 classifiers were fitted. More details can be found in \citet{ariely2025reflective}.   

\item \textbf{Few-shot LLMs}:
The generative models -- GPT-$4$o, GPT-$5.2$ , and Claude Opus 4.5 -- were prompted with assessment instructions. Each category in the rubric had its own prompt for each item, overall $2\times 10 = 20$ prompts. The prompts followed a straightforward few-shot ASAS strategy and their general structure was: role definition, task definition, per-category scoring instructions, and few-shot examples, split equally between positive and negative examples. An example of a prompt for one of the items and categories is provided in Appendix \ref{apd: prompt_example}. We acknowledge that there are various prompting strategies for ASAS. Our approach was to rely on established prompting strategy developed and tested in previous ASAS research. Thus, we adopted the prompts of \citet{ariely2025reflective}, which were initially developed for a GPT-$4$o grader.  We note that \citet{ariely2025reflective} tried several combinations of few-shot examples, specifically $2/4/6/8/10$-shot (with an equal split between positive and negative examples). 

\textbf{\textit{Determining the Number of Few-Shot Examples for the Frontier Models}}. Prior to experimenting with the costly models (GPT-$5.2$ and Claude Opus 4.5), we analyzed the results of running the $2/4/6/8/10$-shot prompts with GPT-$4$o. The results  showed that 10-shot prompting achieved better results, both in terms of overall (average) agreement and in terms of mid-range fairness, as can be seen in Figure 1. Thus, for the costly models, we ran only the 10-shot version. For the few-shot examples, we randomly chose five positive and five negative examples from the training set. The total cost of running the ten 10-shot prompts on the test set for each item for each of the two LLMs (40 prompts in total) was around \$1,000.

\end{enumerate}


\section{Results}

\begin{figure*}[t]
  \includegraphics[width=0.48\linewidth]{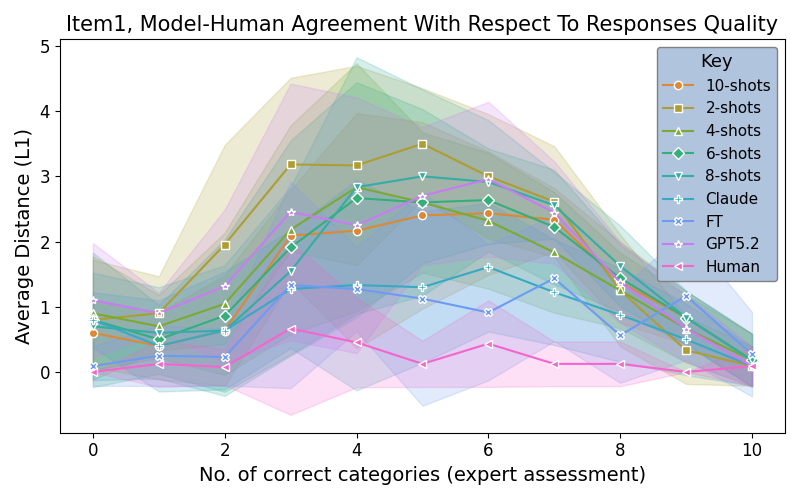} \hfill
  \includegraphics[width=0.48\linewidth]{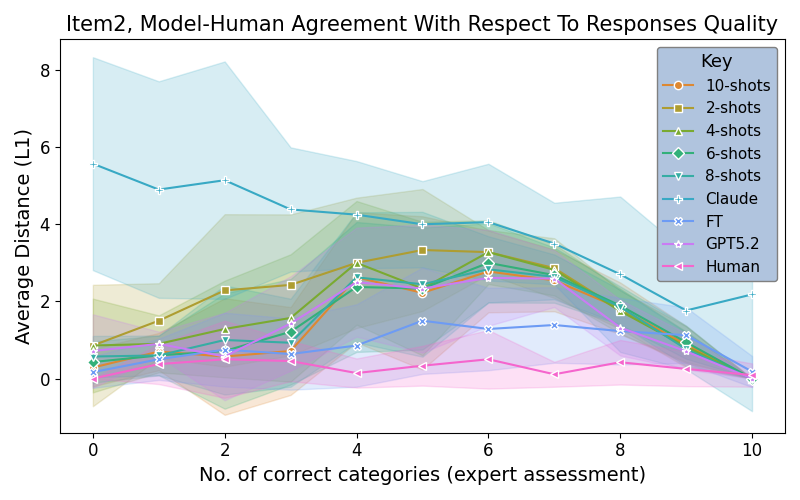}
  \caption {Human-Model Agreement With Respect To Student's Response Quality For All The Models}
  \label{fig:all-shots-gpt-models-vs-humans}
\end{figure*}

Table~\ref{tab:main_results} shows the average L$1$ distance between the gold labels and the scoring of the second human rater (H2) and the AI models, per score level (number of correct categories -- \#CC~-- in the gold labels),
for each of the two items. Figure \ref{fig:all-shots-gpt-models-vs-humans} shows the same information visually, including standard deviations shown in the relevant background color.

Let us first consider the overall performance of the different models on the two items. On average and across score levels, H2 had the highest agreement with the expert, followed by the fine-tuned BERT-based models, followed by the LLMs in few-shot mode. 
Among the GPT models, we did not observe an advantage for GPT5.2 over the prior generation GPT4o model with the same number of few-shot examples (10). For GPT4o -- the only model that was tested with several numbers of few-shot examples -- providing 6-10 examples resulted in approximately similar performance, while providing only four examples resulted in performance degradation on Item 2, and providing only two examples resulted in a substantial degradation in the middle range of the scores -- difference of 3 categories or more, on average, for scores with 4-6 correct categories out of 10 -- for both items. 

Analyzing the results by score level (number of correct categories out of 10), we observe excellent AI-model performance on the extremes -- best and worst quality responses -- regardless of model-type and amount of task-specific data. In contrast, there is a degradation for partially correct responses, especially for the few-shot models. 

While on average fine-tuned models performed better than few-shot ones, the GPT models in all few-shot scenarios -- including those with only two examples -- performed very well on fully or almost fully correct answers (score 9-10), showing average distance of less than one category from the gold score. On the fully correct responses, the GPT models outperformed the fine-tuned models on both items. The exceptionally strong performance of few-shot models on the best responses aligns with some similar reports in the literature (e.g., \citet{grevisse2024llm,PhysRevPhysEducRes.19.020163}). 

Next, we observe that all GPT models perform well on the lowest end of the scale -- score levels 0 and 1. With the exception of two-shot GPT4o on Item 2, in all other cases the average distance from the gold label is less than one category. Although fine-tuned models do better, with average distance of at most 0.5 category from the gold label for score 0 and 1 across the two items, few-shot models are also fairly accurate. 




Table~\ref {tab:main_results} shows that all few-shot models suffer a substantial degradation in performance in the middle range of the scores. For all six GPT models,  for each of the four score levels 4-7, for both items, all but one of the 6$\times$4$\times$2 = 48 results show average distance of more than two categories from the gold label. Fine-tuned models also perform worse in this middle range than in the extremes, but the degradation is not as severe -- the maximum average distances from the gold scores is 1.5 categories. In contrast, the human rater H2 does not show a clear pattern of degradation in agreement with the gold labels in the middle of the score range. 

Our final observation is that of the erratic behavior of the Claude model. While the overall and the score-level-dependent patterns of errors are highly consistent across the two items for H2, fine-tuned, and GPT family models, Claude's performance oscillates dramatically -- it is the best few-shot model for Item 1 but is by far the worst model for Item 2.



\begin{table*}[h]
    \centering
    \begin{tabular}{c|rrrrrrrrr|r} \hline
    \#CC & H2&FT&Claude&GPT&GPT&GPT&GPT&GPT&GPT&Me-\\ 
    &&&&5.2&4o10&4o8&4o6&4o4&4o2&dian \\ \hline
    \multicolumn{10}{c}{\textbf{Item 1}}\\ \hline
    0&   0& 0.09&0.80&\cellcolor{gray!25}1.10&0.60&0.70&0.80&0.90&0.80&0.80\\
    1& 0.12& 0.25&0.40&0.90&0.40&0.60&0.50&0.70&0.90&0.50\\
    2& 0.08& 0.23&0.64&\cellcolor{gray!25}1.32&0.64&0.64&0.86&\cellcolor{gray!25}1.05&\cellcolor{gray!25}1.95&0.64\\
    3& 0.67&\cellcolor{gray!25}1.33&\cellcolor{gray!25}1.27&\cellcolor{gray!50}2.45&\cellcolor{gray!50}2.09&\cellcolor{gray!25}1.55&\cellcolor{gray!25}1.91&\cellcolor{gray!50}2.18&\cellcolor{gray!70}3.18&\cellcolor{gray!25}1.91\\
    4&0.45 &\cellcolor{gray!25}1.27&\cellcolor{gray!25}1.33&\cellcolor{gray!50}2.25&\cellcolor{gray!50}2.17&\cellcolor{gray!50}2.83&\cellcolor{gray!50}2.67&\cellcolor{gray!50}2.83&\cellcolor{gray!70}3.17&\cellcolor{gray!50}2.25\\
    5&0.12 &\cellcolor{gray!25}1.12&\cellcolor{gray!25}1.30&\cellcolor{gray!50}2.70&\cellcolor{gray!50}2.4&\cellcolor{gray!70}3.00&\cellcolor{gray!50}2.60&\cellcolor{gray!50}2.60&\cellcolor{gray!70}3.50&\cellcolor{gray!50}2.60\\
    6&0.43 &0.91&\cellcolor{gray!25}1.61&\cellcolor{gray!50}2.96&\cellcolor{gray!50}2.43&\cellcolor{gray!50}2.91&\cellcolor{gray!50}2.64&\cellcolor{gray!50}2.32&\cellcolor{gray!70}3.00&\cellcolor{gray!50}2.43\\
    7&0.12 & \cellcolor{gray!25}1.44&\cellcolor{gray!25}1.22&\cellcolor{gray!50}2.44&\cellcolor{gray!50}2.33&\cellcolor{gray!50}2.56&\cellcolor{gray!50}2.22&\cellcolor{gray!25}1.83&\cellcolor{gray!50}2.61&\cellcolor{gray!50}2.22\\
    8&0.12&0.56&0.88&\cellcolor{gray!25}1.38&\cellcolor{gray!25}1.38&\cellcolor{gray!25}1.62& \cellcolor{gray!25}1.44&\cellcolor{gray!25}1.25&\cellcolor{gray!25}1.25&\cellcolor{gray!25}1.25\\
    9&0&\cellcolor{gray!25}1.17&0.50&0.67&0.83&0.83&0.83&0.67&0.33&0.67\\
    10&0.09&0.27&0.09&0.09&0.18&0.18&0.18&0.18&0.09&0.18\\\hline
    Avg.&0.20&0.79&0.91&\cellcolor{gray!25}1.66&\cellcolor{gray!25}1.40&\cellcolor{gray!25}1.58&\cellcolor{gray!25}1.51&\cellcolor{gray!25}1.50&\cellcolor{gray!25}1.89 \\
    Stdev.&0.42&0.86&0.82&0.98&0.79&0.91&0.96&0.95&0.93 \\\hline
    \multicolumn{10}{c}{\textbf{Item 2}}\\ \hline
0&0   &0.17&\cellcolor{gray!70}5.57&0.71&0.29&0.57&0.43&0.86&0.86&0.57\\
1&0.38&0.50&\cellcolor{gray!70}4.90&0.90&0.70&0.60&0.60&0.90&\cellcolor{gray!25}1.50&0.70\\
2&0.50&0.75&\cellcolor{gray!70}5.14&0.57&0.57&\cellcolor{gray!25}1.00&0.71&\cellcolor{gray!25}1.29&\cellcolor{gray!50}\cellcolor{gray!50}2.29&0.75\\
3&0.45&0.64&\cellcolor{gray!70}4.38&\cellcolor{gray!25}1.43&0.71&0.93&\cellcolor{gray!25}1.21&\cellcolor{gray!25}1.57&\cellcolor{gray!50}2.43&\cellcolor{gray!25}1.21\\
4&0.14&0.86&\cellcolor{gray!70}4.25&\cellcolor{gray!50}2.50&\cellcolor{gray!50}2.62&\cellcolor{gray!50}2.62&\cellcolor{gray!50}2.38&\cellcolor{gray!70}3.00&\cellcolor{gray!70}3.00&\cellcolor{gray!50}2.62\\
5&0.33&\cellcolor{gray!25}1.50&\cellcolor{gray!70}4.00&\cellcolor{gray!50}2.33&\cellcolor{gray!50}2.22&\cellcolor{gray!50}2.44&\cellcolor{gray!50}2.33&\cellcolor{gray!50}2.33&\cellcolor{gray!70}3.33&\cellcolor{gray!50}2.33\\
6&0.50&\cellcolor{gray!25}1.29&\cellcolor{gray!70}4.06&\cellcolor{gray!50}2.61&\cellcolor{gray!50}2.78&\cellcolor{gray!50}2.83&\cellcolor{gray!70}3.00&\cellcolor{gray!70}3.28&\cellcolor{gray!70}3.28&\cellcolor{gray!50}2.83\\
7&0.11&\cellcolor{gray!25}1.39&\cellcolor{gray!70}3.50&\cellcolor{gray!50}2.59&\cellcolor{gray!50}2.55&\cellcolor{gray!50}2.64&\cellcolor{gray!50}2.68&\cellcolor{gray!50}2.82&\cellcolor{gray!50}2.86&\cellcolor{gray!50}2.64\\
8&0.42&\cellcolor{gray!25}1.23&\cellcolor{gray!50}2.70&\cellcolor{gray!25}1.32&\cellcolor{gray!25}1.82&\cellcolor{gray!25}1.86&\cellcolor{gray!25}1.89&\cellcolor{gray!25}1.75&\cellcolor{gray!25}1.79&\cellcolor{gray!50}1.79\\
9&0.25&\cellcolor{gray!25}1.12&\cellcolor{gray!25}1.76&0.71&0.88&0.76&0.94&0.82&0.82&0.82\\
10&0.09&\cellcolor{gray!25}1.18&\cellcolor{gray!50}2.18&0&0&0&0&0&0&0\\ \hline
Avg.&0.29&0.88&\cellcolor{gray!70}3.86&\cellcolor{gray!25}1.42&\cellcolor{gray!25}1.38 &\cellcolor{gray!25}1.48&\cellcolor{gray!25}1.47&\cellcolor{gray!25}1.69&\cellcolor{gray!50}2.01\\ 
Stdev.&0.49&0.85&1.99&0.90&0.94&0.86&0.90&0.97&1.09 \\\hline
    \end{tabular}
    \caption{Average L1 distance per number of correct categories (\#CC), based on the gold labels. H2 is the second human rater.
    Top panel: Item 1; bottom panel: Item 2. Light gray cells show L1 distance $\geq$ 1; darker gray shows L1 distance $\geq$ 2; the darkest gray shows L1 distance $\geq$ 3.}
    \label{tab:main_results}
\end{table*}

\section{Discussion \& Conclusions}

\noindent\textbf{\textit{Quality-conditioned agreement}}. The findings show that agreement is strongly conditioned by response quality and that this relationship differs considerably between human-human and human-AI pairings. Human raters exhibit the highest agreement, which is also stable across the response quality range and shows no mid-range degradation, suggesting that human experts are relatively consistent in interpreting partially correct responses. In contrast, the AI models showed  a  U-shaped pattern:\footnote{Specifically, the graphs in Figure 1 show an inverted U-shape since they show discrepancy rather than agreement.} Agreement is highest at the extremes (fully correct or incorrect responses) and substantially lower in the middle range, indicating that model–human agreement is not uniform across the scoring range. This mid-range degradation pattern complements a recent ASAS study showing that criterion-level agreement declines as cognitive complexity increases \citep{emirtekin2026automatic}, and a similar trend found in project-based learning \citep{usher2025grades}.



\noindent\textbf{\textit{Task Specific Adaptation.}} The results also showed that task-specific adaptation emerged as a key factor in improving agreement. The fine-tuned model, which had access to the highest amount of task specific data, outperformed scoring based on LLMs with few-shot examples across both items, most notably on mid-range responses.  In line with this, the initial experiment on the number of few-shot examples on GPT-$4$o demonstrated a similar pattern, with the U-shape disagreement curve having a larger ‘belly’ for lower numbers of few-shot examples (see Figure 1). While the advantage of fine-tuned models on LLMs in few-shot mode on ASAS tasks  was shown previously on the aggregated level \citep{kortemeyer2024performance, henkel2024can, ferreira2025automatic}, the present work suggests that this advantage may be mostly concentrated in the mid-range examples. In fact, for the fully correct/incorrect responses, LLMs performed well with very little task-specific data.


\noindent\textbf{\textit{Possible explanations for the observed pattern.}} In trying to explain these results, it may be that extreme responses (fully correct or incorrect) provide clearer signals that are easier to classify, while mid-quality responses require nuanced judgment and interpretation of rubric criteria. Experienced human experts may be better at applying such nuanced educational judgment \cite{yacobson2026chemxai}, and encoder models can learn to distinguish between them using labeled data that changes their internal representation in a way that can separate between them. LLMs, when provided with a rubric and few-shot examples, utilize in-context pattern matching to identify semantic and structural alignments between the criteria, the examples, and the student's input. Our empirical results that show that LLMs are less good than fine-tuned encoders on ASAS tasks are in line with previous findings (e.g., \citet{ferreira2025automatic}).

\noindent\textbf{\textit{Implication for automated scoring in practice.}} The great promise of few-shot automated scoring is fast deployment of the automated scoring model, thus saving instructors valuable time through automated evaluation of student responses without having to  wait until sufficient human-scored data is available for fine-tuning. Our results suggest that such scoring may be inequitable, with substantially inaccurate scores concentrating on the partially incorrect responses. However, the high agreement on fully correct and fully incorrect responses, even for LLM models with very few examples, suggests that automated scoring as a first pass can identify these responses and handle them without human intervention, while responses predicted to be in the middle range would be routed to the human rater for scoring. Gradually, as more human-scored data is accumulated and fine-tuning becomes feasible, the scoring can be automated more fully.

\noindent\textbf{\textit{Implications for fairness.}} 
Bias is a well-recognized hazard in automated scoring of constructed responses \cite{madnani-etal-2017-building,loukina2019many, Gorgun2024AlgBias}. Much of the existing literature has focused on demographic-group–based biases, where students from different groups (e.g., by socioeconomic status \cite{chinta2024fairaied, baker2022algorithmic} gender \cite{madnani-etal-2017-building}, or linguistic background \cite{loukina2019many}) receive systematically different scores despite comparable skill levels. However, bias can also arise at the individual level, where similarly skilled students are treated differently due to consistent features of their responses – for example, students who write in a more concise style may be systematically scored differently \cite{madnani-etal-2017-building,klebanov2022automated}. In addition, statistical biases may emerge from properties of the data, for instance, if  examples that include certain properties, such as specific misconceptions, are not represented enough in the data for the models to learn their linguistic signature. 



Statistical bias can also take the form of a representation bias. For example, \citet{gurin2025uncovering} identified what they termed an `Anna Karenina principle', showing that as responses become more incorrect, their representations in the embedding space become more spread, making it harder to automatically cluster  together responses of the same error type. 
This creates a differential measurement error that disproportionately affects lower-quality responses, and may constitute a fairness problem if such clusters are used to design feedback targeted at each knowledge group, as proposed by \citet{ariely2024causal}. In this case, low-quality responses may receive unsuitable feedback.

Our results contribute to this line of research by identifying a form of measurement bias that appears in ASAS. We observe a U-shaped pattern in which scoring errors concentrate in mid-range responses. These responses are typically produced by students whose knowledge is still developing. Such students may benefit the most from formative feedback, as they are often better able to use feedback to refine their understanding than lower-performing students \citep{roll2014benefits,shute2008focus}. Additionally, because mid-range responses contain both correct and incorrect elements, scoring errors~–~especially misjudging correct components – may be particularly confusing. As a result, reduced accuracy in this range may hinder learning at a critical stage, making this a fairness concern.

We define quality-conditioned fairness as the requirement that automated scoring systems maintain consistent accuracy across the full spectrum of response quality. As we showed, this form of fairness is closely tied to the availability of task-specific data, with greater adaptation reducing mid-range errors. By identifying and characterizing this pattern, we turn it from an “unknown bias” into a “known bias,” which is a necessary step toward improving fairness in educational AI systems in general \citep{baker2022algorithmic}.


\noindent\textbf{\textit{Future work.}} 
To better understand the phenomenon of mid-range degradation, it would be useful to extend this research to additional instruments and domains, and to examine the direction of misalignment, as prior work suggests that LLMs may tend to over-score incorrect responses. To improve mid-range alignment, it would be valuable to investigate whether including more mid-quality examples in few-shot prompts can reduce the observed degradation in scoring. A more human-in-the-loop approach worth investigating is whether task-specific prompts can achieve higher alignment, and how to train domain experts or teachers to develop and validate them within their local context. 

\noindent\textbf{\textit{Contribution.}} This work identifies mid-range scoring degradation in ASAS and shows that it is strongly governed by the degree of task-specific adaptation. A key takeaway is that LLM-based ASAS systems must be evaluated for quality-conditioned agreement, with particular attention to mid-range responses.

\section*{Limitations} 
The prompting approach used for the LLM-based scoring mechanisms followed a common few-shot strategy drawn from prior work. However, these prompts were originally designed and validated for GPT-$4$o. It is possible that alternative prompting strategies, or prompts tailored to each specific model, could yield different performance patterns. 
Additionally, the assessment used in this study consisted of two items in biology. Although these items were selected to represent common question types found in standard assessments on the topic, they still constitute a limited sample of possible tasks. Thus, the generalizability of the findings to other items and domains remains to be established.

This research does not employ standard evaluation metrics such as Quadratic Weighted Kappa (QWK) or Pearson’s correlation, which may hinder direct comparison with existing ASAS benchmarks. Incorporating standard quality-conditioned agreement metrics should be considered in future work.

\section*{Ethics statement}
We acknowledge that this work complies with the ACL Code of Ethics. The research and its data collection procedures were approved by the Institutional Review Board and the Ministry of Education.

\section*{Acknowledgments}
The authors thank Cipy Hofman for her contribution to the research. This work was supported by the Knell Family Institute for Artificial Intelligence, Israel.

\bibliography{custom}

@article{ariely2025reflective,
  author={Ariely, Moriah and Salman, Asaf and Yarden, Anat and Alexandron, Giora},
title={Reflective prompt engineering: a new strategy for automated short answer scoring in biology},
journal={International Journal of Science Education},
  pages={1--23},
  year={2025}
}

@article{wu2025unveiling,
  title   = {Unveiling Scoring Processes: Dissecting the Differences Between {LLM}s and Human Graders in Automatic Scoring},
  author  = {Wu, X. and Saraf, P. P. and Lee, G. and others},
  journal = {Technology, Knowledge and Learning},
  year    = {2025},
  doi     = {10.1007/s10758-025-09836-8},
  url     = {https://doi.org/10.1007/s10758-025-09836-8}
}

@article{PhysRevPhysEducRes.19.020163,
  title = {Toward {AI} grading of student problem solutions in introductory physics: A feasibility study},
  author = {Kortemeyer, Gerd},
  journal = {Phys. Rev. Phys. Educ. Res.},
  volume = {19},
  issue = {2},
  pages = {020163},
  numpages = {20},
  year = {2023},
  month = {Nov},
  publisher = {American Physical Society},
  doi = {10.1103/PhysRevPhysEducRes.19.020163},
  url = {https://link.aps.org/doi/10.1103/PhysRevPhysEducRes.19.020163}
}

@article{grevisse2024llm,
  title={L{L}{M}-based automatic short answer grading in undergraduate medical education},
  author={Gr{\'e}visse, Christian},
  journal={BMC Medical Education},
  volume={24},
  number={1},
  pages={1060},
  year={2024},
  publisher={Springer}
}

@inproceedings{chang2024automatic,
  title={Automatic short answer grading for finnish with chatgpt},
  author={Chang, Li-Hsin and Ginter, Filip},
  booktitle={Proceedings of the AAAI Conference on Artificial Intelligence},
  volume={38},
  number={21},
  pages={23173--23181},
  year={2024}
}

@article{ariely2024causal,
  title={Causal-mechanical explanations in biology: Applying automated assessment for personalized learning in the science classroom},
  author={Ariely, Moriah and Nazaretsky, Tanya and Alexandron, Giora},
  journal={Journal of Research in Science Teaching},
  volume={61},
  number={8},
  pages={1858--1889},
  year={2024},
  publisher={Wiley Online Library}
}

@article{usher2025grades,
  title={Who grades best? Comparing ChatGPT, peer, and instructor evaluations across varying levels of student project quality},
  author={Usher, Maya and Faraon, Montathar},
  journal={Assessment \& Evaluation in Higher Education},
  pages={1--20},
  year={2025},
  publisher={Taylor \& Francis}
}

@article{roll2014benefits,
  title={On the benefits of seeking (and avoiding) help in online problem-solving environments},
  author={Roll, Ido and Baker, Ryan SJ d and Aleven, Vincent and Koedinger, Kenneth R},
  journal={Journal of the Learning Sciences},
  volume={23},
  number={4},
  pages={537--560},
  year={2014},
  publisher={Taylor \& Francis}
}

@article{shute2008focus,
  title={Focus on formative feedback},
  author={Shute, Valerie J},
  journal={Review of educational research},
  volume={78},
  number={1},
  pages={153--189},
  year={2008},
  publisher={Sage Publications}
}

@inproceedings{henkel2024can,
author = {Henkel, Owen and Hills, Libby and Boxer, Adam and Roberts, Bill and Levonian, Zach},
title = {Can Large Language Models Make the Grade? An Empirical Study Evaluating {L}{L}{M}s Ability To Mark Short Answer Questions in K-12 Education},
year = {2024},
isbn = {9798400706332},
publisher = {Association for Computing Machinery},
address = {New York, NY, USA},
url = {https://doi.org/10.1145/3657604.3664693},
doi = {10.1145/3657604.3664693},
booktitle = {Proceedings of the Eleventh ACM Conference on Learning @ Scale},
pages = {300–304},
numpages = {5},
location = {Atlanta, GA, USA},
series = {L@S '24}
}

@article{berti2025emergent,
  title={Emergent abilities in large language models: A survey},
  author={Berti, Leonardo and Giorgi, Flavio and Kasneci, Gjergji},
  journal={arXiv preprint arXiv:2503.05788},
  year={2025}
}

@article{chinta2024fairaied,
  title={FairAIED: Navigating fairness, bias, and ethics in educational {AI} applications},
  author={Chinta, Sribala Vidyadhari and Wang, Zichong and Yin, Zhipeng and Hoang, Nhat and Gonzalez, Matthew and Quy, T Le and Zhang, Wenbin},
  journal={arXiv preprint arXiv:2407.18745},
  year={2024}
}

@inproceedings{madnani-etal-2017-building,
    title = "Building Better Open-Source Tools to Support Fairness in Automated Scoring",
    author = "Madnani, Nitin  and
      Loukina, Anastassia  and
      von Davier, Alina  and
      Burstein, Jill  and
      Cahill, Aoife",
    editor = "Hovy, Dirk  and
      Spruit, Shannon  and
      Mitchell, Margaret  and
      Bender, Emily M.  and
      Strube, Michael  and
      Wallach, Hanna",
    booktitle = "Proceedings of the First {ACL} Workshop on Ethics in Natural Language Processing",
    month = apr,
    year = "2017",
    address = "Valencia, Spain",
    publisher = "Association for Computational Linguistics",
    url = "https://aclanthology.org/W17-1605/",
    doi = "10.18653/v1/W17-1605",
    pages = "41--52"
}

@inproceedings{ding-etal-2020-dont,
    title = "Don{'}t take ``nswvtnvakgxpm'' for an answer {--}The surprising vulnerability of automatic content scoring systems to adversarial input",
    author = "Ding, Yuning  and
      Riordan, Brian  and
      Horbach, Andrea  and
      Cahill, Aoife  and
      Zesch, Torsten",
    editor = "Scott, Donia  and
      Bel, Nuria  and
      Zong, Chengqing",
    booktitle = "Proceedings of the 28th International Conference on Computational Linguistics",
    month = dec,
    year = "2020",
    address = "Barcelona, Spain (Online)",
    publisher = "International Committee on Computational Linguistics",
    url = "https://aclanthology.org/2020.coling-main.76/",
    doi = "10.18653/v1/2020.coling-main.76",
    pages = "882--892"
}

@article{emirtekin2026automatic,
  title={Automatic Short-Answer Grading in Sustainability Education: {AI}--Human Agreement},
  author={Emirtekin, Emrah and {\"O}zarslan, Yasin},
  journal={Journal of Computer Assisted Learning},
  volume={42},
  number={1},
  pages={e70160},
  year={2026},
  publisher={Wiley Online Library}
}

@article{haller2022survey,
  title={Survey on automated short answer grading with deep learning: from word embeddings to transformers},
  author={Haller, Stefan and Aldea, Adina and Seifert, Christin and Strisciuglio, Nicola},
  journal={arXiv preprint arXiv:2204.03503},
  year={2022}
}

@inproceedings{bonthu2021automated,
  title={Automated short answer grading using deep learning: A survey},
  author={Bonthu, Sridevi and Rama Sree, S and Krishna Prasad, MHM},
  booktitle={International cross-domain conference for machine learning and knowledge extraction},
  pages={61--78},
  year={2021},
  organization={Springer}
}

@inproceedings{chen2004marriage,
  title={On the marriage of lp-norms and edit distance},
  author={Chen, Lei and Ng, Raymond},
  booktitle={Proceedings of the Thirtieth international conference on Very large data bases-Volume 30},
  pages={792--803},
  year={2004}
}

@article{liu2019roberta,
  title={Roberta: A robustly optimized bert pretraining approach},
  author={Liu, Yinhan and Ott, Myle and Goyal, Naman and Du, Jingfei and Joshi, Mandar and Chen, Danqi and Levy, Omer and Lewis, Mike and Zettlemoyer, Luke and Stoyanov, Veselin},
  journal={arXiv preprint arXiv:1907.11692},
  year={2019}
}

@techreport{anthropic2025,
  title       = {Claude Opus 4.5 System Card},
  author      = {{Anthropic}},
  institution = {Anthropic},
  year        = {2025},
  month       = nov,
  note        = {System card},
  url         = {https://www.anthropic.com/claude-opus-4-5-system-card}
}

@article{achiam2023gpt,
  title={Gpt-4 technical report},
  author={ OpenAI},
  journal={arXiv preprint arXiv:2303.08774},
  year={2023}
}

@article{Aaditya2025gpt,
  title={Open{AI} gpt-5 system card},
  author={Singh, Aaditya and Fry, Adam and Perelman, Adam and Tart, Adam and Ganesh, Adi and El-Kishky, Ahmed and McLaughlin, Aidan and Low, Aiden and Ostrow, AJ and Ananthram, Akhila and others},
  journal={arXiv preprint arXiv:2601.03267},
  year={2025}
}

@inproceedings{schleifer2023transformer,
  title={Transformer-based Hebrew NLP models for short answer scoring in biology},
  author={Gurin Schleifer, Abigail and Klebanov, Beata Beigman and Ariely, Moriah and Alexandron, Giora},
  booktitle={Proceedings of the 18th Workshop on Innovative Use of NLP for Building Educational Applications (BEA 2023)},
  pages={550--555},
  year={2023}
}

@inproceedings{bexte2023similarity,
  title={Similarity-based content scoring-a more classroom-suitable alternative to instance-based scoring?},
  author={Bexte, Marie and Horbach, Andrea and Zesch, Torsten},
  booktitle={Findings of the association for computational linguistics: Acl 2023},
  pages={1892--1903},
  year={2023}
}

@inproceedings{ferreira2025automatic,
author = {Ferreira Mello, Rafael and Pereira Junior, Cleon and Rodrigues, Luiz and Pereira, Filipe Dwan and Cabral, Luciano and Costa, Newarney and Ramalho, Geber and Gasevic, Dragan},
title = {Automatic Short Answer Grading in the {LLM} Era: Does {GPT}-4 with Prompt Engineering beat Traditional Models?},
year = {2025},
isbn = {9798400707018},
publisher = {Association for Computing Machinery},
url = {https://doi.org/10.1145/3706468.3706481},
doi = {10.1145/3706468.3706481},
booktitle = {Proceedings of the 15th International Learning Analytics and Knowledge Conference},
pages = {93–103},
numpages = {11},
location = {
},
series = {LAK '25}
}

@inproceedings{devlin2018bert,
    title = "{BERT}: Pre-training of Deep Bidirectional Transformers for Language Understanding",
    author = "Devlin, Jacob  and
      Chang, Ming-Wei  and
      Lee, Kenton  and
      Toutanova, Kristina",
    editor = "Burstein, Jill  and
      Doran, Christy  and
      Solorio, Thamar",
    booktitle = "Proceedings of the 2019 Conference of the North {A}merican Chapter of the Association for Computational Linguistics: Human Language Technologies, Volume 1 (Long and Short Papers)",
    month = jun,
    year = "2019",
    address = "Minneapolis, Minnesota",
    publisher = "Association for Computational Linguistics",
    url = "https://aclanthology.org/N19-1423/",
    doi = "10.18653/v1/N19-1423",
    pages = "4171--4186"
}

@inproceedings{yacobson2026chemxai,
  title={Human Experts vs. {LLM}s: Who Is Better at Explaining Student Clustering?},
  author={Yacobson, E. and Rapp, S. and Blonder, R. and Alexandron, G.},
  booktitle={Proceedings of the 2nd Human-Centric eXplainable AI in Education (HEXED) Workshop at EDM 2025},
  year={2025},
  url={https://doi.org/10.35542/osf.io/4ewfg_v1}
}

@article{kortemeyer2024performance,
  title={Performance of the pre-trained large language model GPT-4 on automated short answer grading},
  author={Kortemeyer, Gerd},
  journal={Discover Artificial Intelligence},
  volume={4},
  number={1},
  pages={47},
  year={2024},
  publisher={Springer}
}

@inproceedings{condor2020exploring,
  title={Exploring automatic short answer grading as a tool to assist in human rating},
  author={Condor, Aubrey},
  booktitle={International Conference on Artificial Intelligence in Education},
  pages={74--79},
  year={2020},
  organization={Springer}
}

@inproceedings{chamieh-etal-2024-llms,
    title = "{LLM}s in Short Answer Scoring: Limitations and Promise of Zero-Shot and Few-Shot Approaches",
    author = "Chamieh, Imran  and
      Zesch, Torsten  and
      Giebermann, Klaus",
    editor = {Kochmar, Ekaterina  and
      Bexte, Marie  and
      Burstein, Jill  and
      Horbach, Andrea  and
      Laarmann-Quante, Ronja  and
      Tack, Ana{\"i}s  and
      Yaneva, Victoria  and
      Yuan, Zheng},
    booktitle = "Proceedings of the 19th Workshop on Innovative Use of NLP for Building Educational Applications (BEA 2024)",
    month = jun,
    year = "2024",
    address = "Mexico City, Mexico",
    publisher = "Association for Computational Linguistics",
    url = "https://aclanthology.org/2024.bea-1.25/",
    pages = "309--315"
}

@book{klebanov2022automated,
  title={Automated essay scoring},
  author={Beigman Klebanov, Beata  and Madnani, Nitin},
  year={2022},
  publisher={Springer Nature}
}

@inproceedings{loukina2019many,
  title={The many dimensions of algorithmic fairness in educational applications},
  author={Loukina, Anastassia and Madnani, Nitin and Zechner, Klaus},
  booktitle={Proceedings of the fourteenth workshop on innovative use of NLP for building educational applications},
  pages={1--10},
  year={2019}
}

@incollection{madaio2022beyond,
  title={Beyond “fairness”: Structural (in) justice lenses on {AI} for education},
  author={Madaio, Michael and Blodgett, Su Lin and Mayfield, Elijah and Dixon-Rom{\'a}n, Ezekiel},
  booktitle={The ethics of artificial intelligence in education},
  pages={203--239},
  year={2022},
  publisher={Routledge}
}

@article{baker2022algorithmic,
  title={Algorithmic bias in education},
  author={Baker, Ryan  and Hawn, Aaron},
  journal={International journal of artificial intelligence in education},
  volume={32},
  number={4},
  pages={1052--1092},
  year={2022},
  publisher={Springer}
}

@article{gurin2025uncovering,
  title={Uncovering measurement biases in {L}{L}{M} embedding spaces: The Anna Karenina Principle and its implications for automated feedback},
  author={Gurin Schleifer, Abigail and Beigman Klebanov, Beata and Alexandron, Giora},
  journal={International Journal of Artificial Intelligence in Education},
  volume={35},
  number={5},
  pages={2821--2855},
  year={2025},
  publisher={Springer}
}

@article{shmidman2023dictabert,
  title={Dictabert: A state-of-the-art bert suite for modern hebrew},
  author={Shmidman, Shaltiel and Shmidman, Avi and Koppel, Moshe},
  journal={arXiv preprint arXiv:2308.16687},
  year={2023}
}

@article{camilli2006test,
  title={Test fairness},
  author={Camilli, Gregory},
  journal={Educational measurement},
  volume={4},
  pages={221--256},
  year={2006}
}

@incollection{johnson2023evaluating,
  title={Evaluating fairness of automated scoring in educational measurement},
  author={Johnson, Matthew S and McCaffrey, Daniel F},
  booktitle={Advancing natural language processing in educational assessment},
  pages={142--164},
  year={2023},
  publisher={Routledge}
}

@article{xi2010we,
  title={How do we go about investigating test fairness?},
  author={Xi, Xiaoming},
  journal={Language testing},
  volume={27},
  number={2},
  pages={147--170},
  year={2010},
  publisher={Sage Publications Sage UK: London, England}
}

@article{williamson2012framework,
author = {Williamson, David M. and Xi, Xiaoming and Breyer, F. Jay},
title = {A Framework for Evaluation and Use of Automated Scoring},
journal = {Educational Measurement: Issues and Practice},
volume = {31},
number = {1},
pages = {2-13},
doi = {https://doi.org/10.1111/j.1745-3992.2011.00223.x},
url = {https://onlinelibrary.wiley.com/doi/abs/10.1111/j.1745-3992.2011.00223.x},
eprint = {https://onlinelibrary.wiley.com/doi/pdf/10.1111/j.1745-3992.2011.00223.x},
year = {2012}
}

@article{ariely2022machine,
  title={Machine learning and {H}ebrew {NLP} for automated assessment of open-ended questions in biology},
  author={Ariely, Moriah and Nazaretsky, Tanya and Alexandron, Giora},
  journal={International Journal of Artificial Intelligence in Education},
  volume = {33},
  number = {1},
  pages={1--34},
  year={2023},
  publisher={Springer}
}

@article{lin2023unlocking,
  title={The unlocking spell on base {LLM}s: Rethinking alignment via in-context learning},
  author={Lin, Bill Yuchen and Ravichander, Abhilasha and Lu, Ximing and Dziri, Nouha and Sclar, Melanie and Chandu, Khyathi and Bhagavatula, Chandra and Choi, Yejin},
  journal={arXiv preprint arXiv:2312.01552},
  year={2023}
}

@article{Gorgun2024AlgBias,
author = {Gorgun, Guher and Yildirim-Erbasli, Seyma N.},
title = {Algorithmic Bias in {BERT} for Response Accuracy Prediction: A Case Study for Investigating Population Validity},
journal = {Journal of Educational Measurement},
volume = {63},
number = {1},
pages = {e12420},
doi = {https://doi.org/10.1111/jedm.12420},
url = {https://onlinelibrary.wiley.com/doi/abs/10.1111/jedm.12420},
eprint = {https://onlinelibrary.wiley.com/doi/pdf/10.1111/jedm.12420},
year = {2026}
}

@inproceedings{li2021semantic,
  title={A semantic feature-wise transformation relation network for automatic short answer grading},
  author={Li, Zhaohui and Tomar, Yajur and Passonneau, Rebecca J},
  booktitle={Proceedings of the 2021 Conference on Empirical Methods in Natural Language Processing},
  pages={6030--6040},
  year={2021}
}
\newpage
\appendix

\section{Prompt Example}\label{apd: prompt_example}
The general prompt structure is taken from \citet{ariely2025reflective}. The scoring instructions are tuned per category. Each category-specific prompt is then followed by randomly chosen few-shot positive and negative examples.
\begin{enumerate}
    \item \textbf{Role definition:}\\
    'You are an expert in education and assessment in biology.'
    \item \textbf{Task definition and general instructions:}\\
    'You are required to assign a score to category in a student's response titled: "Changes in the rate or amount of energy Production." Note that although the overall context of the response is important, you should focus solely on the students' reference to changes in the rate or amount of energy or ATP production. Anything else mentioned around this category may be significant for the response, but irrelevant for evaluating this specific category.'
    \item \textbf{Instructions for scoring:}\\
    'If the category is present in the response, score 1. If not, score 0.'
\end{enumerate}

\end{document}